\pdfoutput=1

\documentclass[11pt]{article}

\usepackage[preprint]{acl}

\usepackage{times}
\usepackage{latexsym}

\usepackage[T1]{fontenc}

\usepackage[utf8]{inputenc}

\usepackage{microtype}

\usepackage{inconsolata}

\usepackage{graphicx}
\usepackage{amsthm}
\usepackage{amsmath}
\usepackage{amssymb}
\usepackage{mathtools}
\theoremstyle{plain}

\usepackage{booktabs, array} %
\newcolumntype{H}{>{\setbox0=\hbox\bgroup}c<{\egroup}@{}}
\usepackage[textsize=tiny]{todonotes}
\usepackage{makecell}
\usepackage{colortbl}
\usepackage{xcolor}
\definecolor{ForestGreen}{RGB}{34, 139, 34}

\NewDocumentCommand{\yz}
{ mO{} }{\textcolor{blue}{\textsuperscript{\textit{Yizhe}}\textsf{\textbf{\small[#1]}}}}

\NewDocumentCommand{\rxz}
{ mO{} }{\textcolor{teal}{\textsuperscript{\textit{rxz}}{{\small[#1]}}}}

\NewDocumentCommand{\rb}
{ mO{} }{\textcolor{ForestGreen}{\textsuperscript{\textit{rb}}{{\small[#1]}}}}

\usepackage{hyperref} 
\usepackage{url}            %
\usepackage{amsfonts}       %
\usepackage{nicefrac}       %
\usepackage{wrapfig,lipsum}
\usepackage{tcolorbox}
\usepackage{color}

\usepackage[export]{adjustbox}
\usepackage{multirow}
\usepackage{xspace}
\usepackage{subcaption}
\usepackage{paralist}

\newcommand{\ours}{\textsc{SAGE}\xspace}

\author{%
  Yizhe Zhang, \quad Navdeep Jaitly \\
  Apple\\
  \texttt{ \{yizzhang, njaitly\}@apple.com} \\
}

\title{\ours: Steering Dialog Generation with Future-Aware State-Action Augmentation}

\begin{document}

\maketitle
\begin{abstract}
Recent advances in large language models have demonstrated impressive capabilities in task-oriented applications, yet building emotionally intelligent chatbots that can engage in natural, strategic conversations remains a challenge. We present a novel approach called \ours that uses latent variables to control long-horizon behavior in dialogue generation. At the core of our method is the State-Action Chain (SAC), which augments standard language model fine-tuning by introducing latent variables that encapsulate emotional states and conversational strategies between dialogue turns. During inference, these variables are generated before each response, enabling coarse-grained control over dialogue progression while maintaining natural interaction patterns. We also introduce a self-improvement pipeline that leverages dialogue tree search, LLM-based reward modeling, and targeted fine-tuning to optimize conversational trajectories. Our experimental results show that models trained with this approach demonstrate improved performance in emotional intelligence metrics while maintaining strong capabilities on LLM benchmarks. The discrete nature of our latent variables facilitates search-based strategies and provides a foundation for future applications of reinforcement learning to dialogue systems, where learning can occur at the state level rather than the token level. \url{https://github.com/apple/ml-sage-dialog-gen}

\end{abstract}

\section{Introduction}

\begin{figure*}[ht!]
    \centering
    \includegraphics[width=1.0\linewidth]{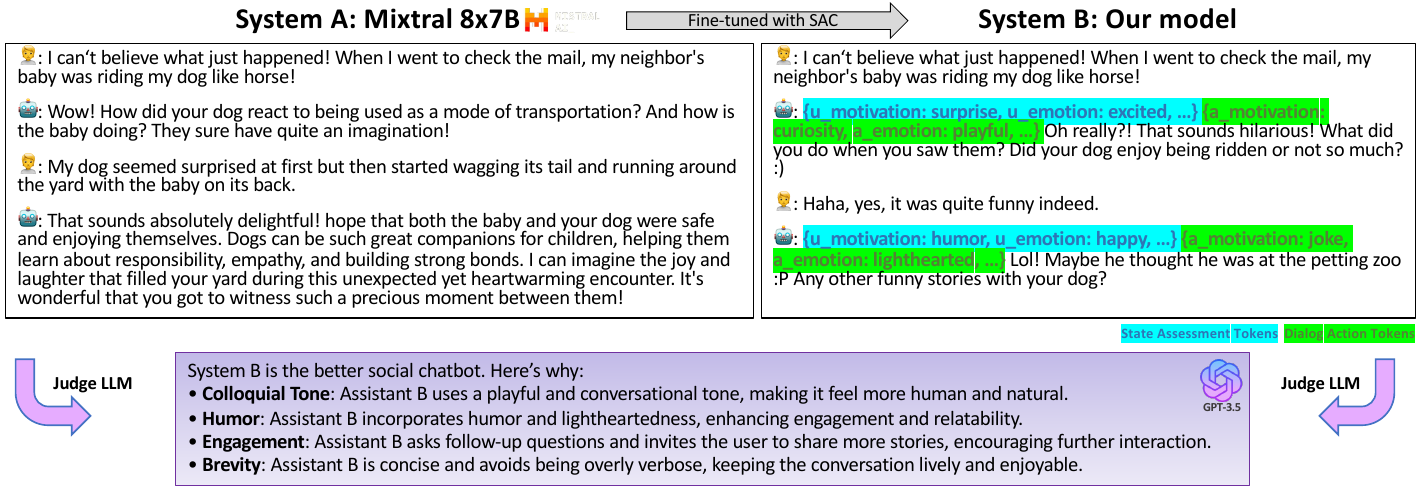}
    \caption{We propose to augment a base-LLM via State-Action Chain (SAC) to provide more control in a multiturn social-oriented dialogue scenario. During inference time, the resulting model first generates additional State Assessment Tokens and Dialog Action Tokens before generating the actual response.}
    \vspace{-2mm}
    \label{fig:comparison}
\end{figure*}
Recent advances in large language models (LLMs) have shown impressive capabilities in task-oriented applications like question-answering, mathematics, and coding \citep{xu2024education, liu2024algorithms, hou2024software}. However, effective human-AI interaction requires not just problem-solving abilities but also emotional intelligence and interpersonal skills. This has led to emotion-based chatbots like Character AI \citep{CharacterAI2023} and Pi \citep{PI2023}, which focus on emotional connection and engagement. These models are designed to understand emotions, provide empathy, and build trust through natural conversations, finding applications in education, healthcare, counseling, and gaming \citep{beale2009affective, reilly1996believable}.

Building effective emotional chatbots presents unique challenges. Unlike task-oriented interactions, emotional conversations require strategic navigation without clear objectives. Therefore, (1) chatbots need to steer conversations \textit{strategically} by planning ahead—an AI therapist must balance various conversational moves to build trust and facilitate disclosure, while an AI negotiator needs to integrate multiple techniques to achieve desired outcomes. (2) The chatbot should be \textit{proactive} and \textit{interactive} by asking clarifying questions and providing customized information based on real-time feedback—a career counselor might probe into a client's experiences to offer personalized guidance, while a fitness coach might assess energy levels to adjust workout recommendations.

These capabilities require effective learning mechanisms for handling sparse rewards across multiple conversation turns. While RLHF \citep{ouyang2022training} and DeepSeek R1 \citep{guo2025deepseek} have succeeded in complex reasoning tasks \citep{weng2023agent, wang2023survey, ahn2024large, wang2024executable, wang2024openhands}, operating directly on the huge token action space with long horizon remains challenging. Previous work \citep{chai2024marlhf} has shown that using \textbf{macro actions} improves credit assignment and learning efficiency.

We propose \ours (\textbf{S}tate \textbf{A}ugmented \textbf{GE}neration) that augments language model fine-tuning with meaningful \textit{latent variables} capturing longer-term conversational information. These model-generated variables help balance short- and long-term predictions by providing explicit representations of dialogue states and actions that encode strategic information beyond immediate token-level generation. By learning to predict these high-level variables first, the model can make more informed decisions about utterance generation that consider both immediate context and long-term conversational goals.

We introduce the \textit{State-Action Chain} (SAC), which extends chain-of-thought approaches to capture dialogue states' evolution. As shown in Figure~\ref{fig:comparison}, SAC maintains abstract representations of emotional states and conversational dynamics, enabling coarse-grained control while maintaining natural interaction. This approach separates high-level planning from language generation, making it suitable for reinforcement learning at the state level rather than token level. SAC is a \textbf{future-looking annotation strategy}, where state and action labels are generated by considering the complete dialogue context rather than individual utterances in isolation, enabling the model to develop strategic thinking capabilities.

We developed a self-improvement pipeline combining data augmentation, evaluation, tree search, and fine-tuning techniques. This pipeline explores alternative conversational paths, uses rejection sampling based on LLM-derived rewards, and retrains using the most effective trajectories. Our results show improved performance while maintaining capabilities on standard benchmarks. We released our dataset and model checkpoints \footnote{\url{https://github.com/apple/ml-sage-dialog-gen}}.

\section{Related Work}
\textbf{Decision Transformer} The Decision Transformer (DT) \citep{chen2021decision} leverages a transformer architecture to model trajectory data as a sequence of states, actions, and rewards, effectively casting decision-making problems as supervised learning tasks. Recent works have applied DT to diverse applications including gaming \citep{chen2021decision}, robotics \citep{janner2021trajectory} and chip design \citep{lai2023chipformer}. In emotional chatbot contexts, DT's ability to model long-term dependencies could be pivotal for balancing immediate conversational moves with long-term goals like trust-building and engagement. Our work takes the initial step by augmenting utterances with states and actions.

\textbf{Latent Variable Approaches in Dialogue Generation} Several works have explored the use of latent variables to enhance dialogue generation. \citet{serban2019piecewise} introduced a hierarchical latent variable model that captures discourse-level structure in conversations, while \citet{bao2019platonic} proposed a dialogue generation model with discrete latent variables to model conversation flow and speaker intentions. Our SAC approach differs by focusing specifically on emotional states and conversational strategies, with a future-looking annotation process that considers the complete dialogue context for more accurate state assessment.

\paragraph{Chain-of-Thought}
Chain-of-thought (CoT) \citep{wei2022chain} has demonstrated remarkable effectiveness in tasks requiring logical and mathematical reasoning. \citet{snell2024scaling} shows that test-time compute scaling can be more efficient and effective than scaling the model parameters. Following this paradigm, our work incorporates CoT-style reasoning into emotional chatbot interactions by decomposing dialogue generation into a high-level, abstract planning stage that represents the evolution of dialogue states and emotional dynamics, and a language realization stage.

\paragraph{Proactive Dialog Systems}
Proactive dialog systems anticipate user needs and guide conversations toward desired outcomes using hierarchical structures and reinforcement learning. Examples include mixed-initiative systems for problem-solving and models for strategic customer interactions \citep{mehri2020usr}. \citet{hong2023zero} used synthetic colloquial data and offline RL to improve LLMs in goal-oriented dialogues. In emotion-based chatbots, our approach aligns with the need for high-level guidance, where the system predicts emotional states and motivations to sustain meaningful conversations.

\paragraph{State Augmentation for Task-Oriented Dialogue} 
Task-oriented dialogue systems traditionally rely on modularized stages of language understanding, state-tracking, dialog policy learning and utterance generation. However, advances in neural architectures have enabled more flexible and robust task completion by leveraging contextual embeddings and pre-trained language models \citep{budzianowski2018multiwoz}. SOLOIST \citep{peng2021soloist} consolidates modular task-oriented dialogue pipelines into a single transformer-based model with state augmentation. Our work also integrates high-level dialogue guidance but additionally incorporates reasoning mechanisms for dialogue actions in the context of emotional chatbots. In contrast to task-oriented systems which search from a finite number of possible states, emotional chatbots focus on open-ended interactions with unbounded state spaces.

\section{State Augmented Generation}
\subsection{Raw Data Preparation}
We use our in-house conversational dataset extracted from Reddit spanning the years 2005 to 2017, following the recipe from DialoGPT \citep{zhang2019dialogpt}. We applied aggressive filtering by selecting only conversations with more than four turns and where the average length of each utterance exceeds 15 words. To filter out inappropriate language and tune up the sentiment in the resulting models, sentiment analysis was performed on each utterance using the \textsc{SentimentIntensityAnalyzer} from NLTK \citep{bird-loper-2004-nltk}, and we retained only the dialogues where all utterances had a sentiment score above 0.4. Additionally, we filtered the dataset to include only dialogues where at least one utterance ends with a question mark, aiming to encourage the trained model to generate questions more frequently. These filtering steps resulted in a total of 181,388 multi-turn training instances.

\subsection{State-Action Chain Augmentation}
Instead of relying on the model to generate an utterance through next token prediction alone, we want the model to acquire the following capabilities:

\begin{compactitem}
    \item \textbf{State tracking}: Estimate the current dialogue history's state.
    \item \textbf{Policy Learning}: Learn a dialog policy to predict the action based on the current state.
    \item \textbf{Utterance Generation}: Generate an utterance to execute the predicted action.
\end{compactitem}

This approach is comparable to conventional task-oriented chatbot systems that perform goal-oriented tasks like restaurant booking, which employ distinct modules for dialogue state tracking (DST), policy learning, and natural language generation (NLG).

Our goal is to construct an end-to-end data-driven solution for a social chatbot, leveraging the strengths of existing LLMs. Drawing inspiration from the Online Decision Transformer \citep{zheng2022online} and CoT \citep{wei2022chain}, we augment the data such that the state tracking and policy learning processes are explicitly integrated into the standard autoregressive LM learning process. Consequently, the model will produce a series of CoT-style ``Thinking'' tokens, predicting the current state and the action to be taken, as illustrated in Figure~\ref{fig:SAC}.  

The raw dialogue data is first annotated using an LLM (see Appendix~\ref{app:annot_prompt}) with access to the complete dialogue context. Each dialog session is seen as a back-and-forth message exchange between a user and a dialog system, with the user starting the conversation and the system responding next. The extra annotation tokens in the user's message are called ``State Assessment Tokens," while those in the system's message are labeled "Dialog Action Tokens." The State Assessment token is then moved to the start of the system's message to complete the SAC augmentation. When fine-tuning the model, the loss is exclusively on the system's message, requiring the model to predict the state (State Assessment Token) based on the previous user utterance, followed by predicting the action (Dialog Action Token) for the current system's turn before finally generating the system's response.

A key insight of our approach is the \textbf{future-looking nature} of the SAC annotation process. Unlike traditional annotation methods that label each utterance in isolation, our annotation strategy takes into account the entire dialogue context to disambiguate intent and motivation. When annotating the motivation for a current utterance, looking at that single utterance alone might not provide sufficient clarity about the speaker's underlying intent. However, by examining the complete dialogue trajectory from the current point to the end, the annotator model can better understand the consequences and utility of each conversational move, leading to higher quality and more accurate state-action annotations.

This future-aware annotation strategy provides several key benefits. First, it helps disambiguate ambiguous utterances by considering their consequences in the broader conversation context. Second, it enables the model to learn strategic thinking patterns, similar to how a Q-function learns to associate state-action pairs with their expected future value. Third, it can potentially allow for more effective reinforcement learning by providing clearer signals about the long-term utility of different conversational strategies.

\begin{figure*}[ht!]
    \centering
    \includegraphics[width=1.0\linewidth]{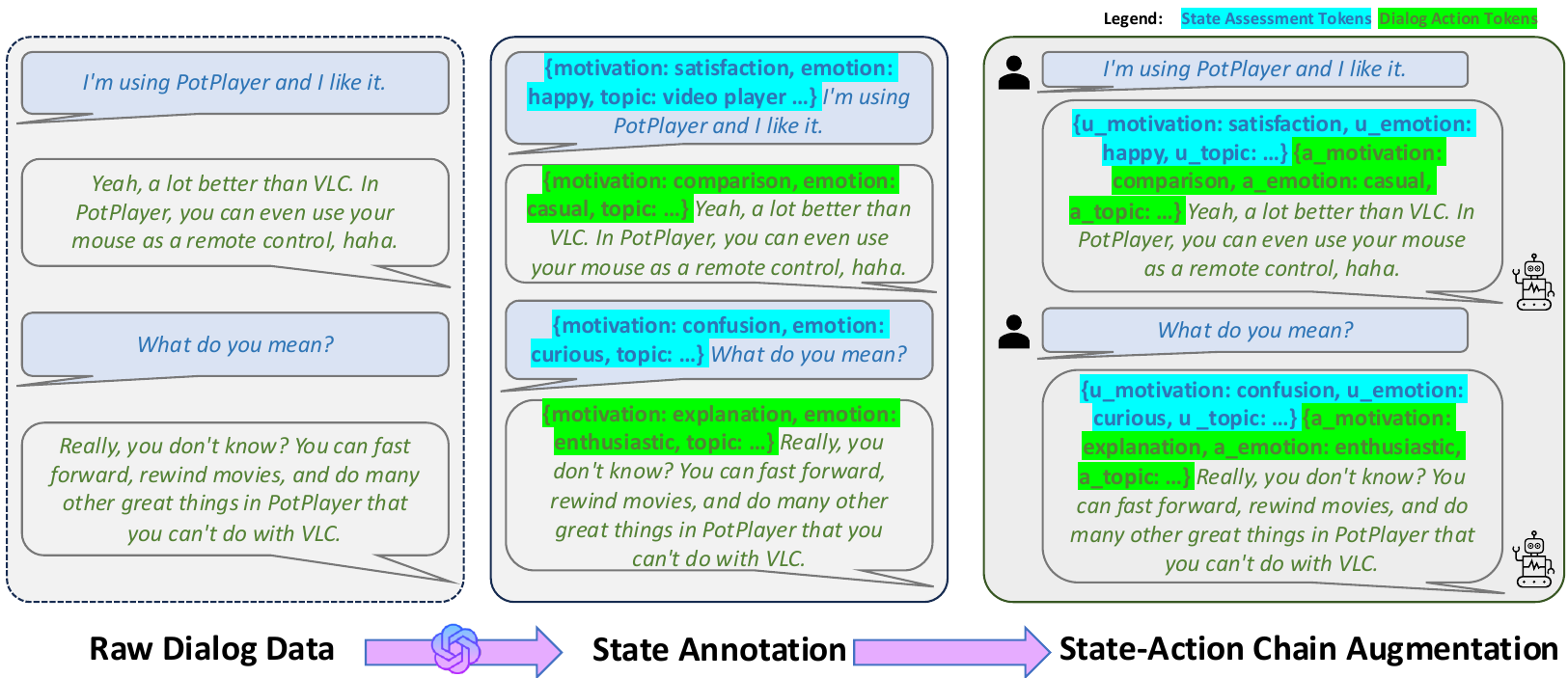}
    \caption{State-Action Chain (SAC) augmentation contains two stages. The first stage annotates the dialog using an LLM. The second stage moves the user's State Assessment Token (highlighted in blue) to the start of the system's message, enabling the system to predict the state based on the user's previous message, followed by predicting the Dialog Action Token (highlighted in green), and finally generating the system's response during fine-tuning with loss focused solely on the system's message.}
    \label{fig:SAC}
\end{figure*}

The advantages of this approach are twofold. First, it enables fine-grained control—the abstract nature of state and action tokens facilitates direct manipulation, allowing reinforcement learning to adjust only a few action tokens rather than the entire model generation. This refinement can significantly enhance the efficacy of long-horizon RL training. Secondly, it enables additional planning and reasoning for generation, akin to CoT.

As shown in Figure~\ref{fig:SAC}, for each dialogue, we initially use an LLM (Mixtral 8x7B) \citep{jiang2024mixtral} to annotate the dialog state by attaching relevant states (e.g., motivation, emotion) to the beginning of each dialog utterance. The annotation process is \textbf{future-aware}, meaning the annotator model has access to the complete dialogue context when labeling each utterance. This allows for more accurate state assessment by considering how each conversational move contributes to the overall dialogue trajectory and desired outcomes. Subsequently, the states from the odd user utterance are amalgamated with those from the even assistant utterance to create assistant responses containing three parts: user state, assistant state, assistant utterance. During generation, the model generates these three parts in sequence, mirroring the \textit{state prediction}, \textit{action prediction}, and \textit{utterance generation} process.

\subsection{Finetuning for State Prediction}
Using the data augmented with the State-Action Chain annotations, we finetuned a model ($\ours_1$) using a Mixtral 8x7B as the base model ($\ours_0$). The model underwent 5 epochs of finetuning. We used LoRA \citep{hu2022lora} instead of dense-training because it enabled the model to learn state generation while preserving the capabilities of the starting model.

To assess the effectiveness of SAC, we trained a baseline model without SAC augmentation using the same number of training iterations and setup. This resulting model is denoted as $\ours_1 (-\text{SAC})$.

\subsection{Iterative Dialog Tree Search and Refinement via Self-Play Rollout}
Starting from the $\ours_1$ model, we perform an iterative search and refinement process based on \textbf{self-play} to enhance its quality. We leverage seed situational scenarios from the EmpatheticDialogs dataset \citep{rashkin2019towards}, which comprises 19,533/2,770/2,547 instances for training, dev, and test sets respectively. Each instance contains a situational statement like ``My friend got tickets to the Superbowl and not me.'' along with its corresponding sentiment, such as ``jealous''. We only use the sentiment to empirically verify the effectiveness of our predicted emotion.

The dialogue tree search process operates as follows: During the $k$-th iteration, the current model $\ours_k$ performs the role of the agent, while $\ours_1$ plays the role of the user. We use each situational statement from the training set as the initial utterance from the user and then prompt both models with the current dialogue history to simulate a conversation between two speakers for up to 12 turns. The generation process involves topK sampling with K = 100, temperature = 1.1, and repetition penalty = 1.1.

For each turn, the agent model $\ours_k$ generates 16 candidate responses using beam search with diverse sampling. Subsequently, an external selector LLM (Mixtral 8x7B) evaluates and selects the best candidate based on predefined properties that an ideal companion chatbot should exhibit, including consistency, humor, sympathy, informativeness, appropriateness, and respect (see box in Section~\ref{sec:eval}). The selection process considers both the immediate quality of the response and its potential to lead to engaging future interactions. On the user side, only one generation is produced per turn to maintain conversation consistency. The resulting high-quality trajectories are used to fine-tune the current model $\ours_k$ to the subsequent model iteration $\ours_{k+1}$ using LoRA, with user utterances masked out during training to focus on improving agent responses. See Figure~\ref{fig:treesearch} in Appendix as an example of the tree search process.

We iterate over this refinement process until $\ours_{k+1}$ reaches a level comparable to $\ours_k$, based on the model evaluation pipeline and metrics detailed in the subsequent section. Through experimentation, we observed that improvements beyond 2 iterations tend to be marginal, suggesting that $\ours_3$ has already reached the saturation point in the search-and-refine phase.

\subsection{Preference Learning}
We then conducted preference learning using DPO \citep{rafailov2024direct} on the $\ours_3$ model. We use the selected utterance from the agent model as a positive example, and select one of the rejected utterances randomly as the negative example. The resulting model is denoted as $\ours_{DPO}$.

\subsection{Model Evaluation}
\label{sec:eval}
We need to quantitatively evaluate the performance differences among various model variants and iterations. Human evaluations can be costly, so we opt for LLM-based assessments, as human-preference aligned LLMs are shown to serve as a cost-effective and dependable alternative to human judgments \citep{zheng2023judging}. We first roll-out dialogues between the tested agent model and a user model (based on $\ours_1$) for up to 16 turns on each instance in the dev set. We then use a Judge LLM for pairwise comparisons between the generated conversations using two models. This involves starting from each situational statement in the dev set. The judge model is then provided two conversation sessions and asked to determine which one is superior using the prompt in Appendix~\ref{app:judge_prompt}.

We refrain from using particular desirable properties as criteria for evaluation and task the judging model with assessing based on its inherent understanding of what makes a good social chatbot. To mitigate the potential bias introduced by the order of the presented dialogues, we conduct two assessments for each pair by switching the positions of dialogues A and B. The judgments are considered reliable only if they remain consistent across both evaluations. Subsequently, we aggregate preferences from all valid votes to determine the ultimate winner model.

\subsection{Inference Time State Manipulation}
Our state prediction facilitates effortless and seamless manipulation of states during runtime. Through small adjustments to one single logit in the agent's generated action during inference, we can conveniently modify aspects such as the desired emotion and motivation we want to apply to the agent, leading to noticeable changes in overall behavior across subsequent interactions. For instance, rather than training a new model for a more humorous response, we can simply augment specific keyword logits like ``humor'' and ``joke'' after the ``a\_motivation'' by a value (e.g., 1.0). This approach empowers us to customize the model's behavior on-the-fly during the inference process.

\section{Results}
\paragraph{LLM-judge based evaluation}
Following~\S\ref{sec:eval}, we compare the various versions of the models, namely $\ours_1, \ours_2, \ours_3$, with two Judge LLMs, namely GPT-3.5 and Mixtral 8x7B. The LLM selector in the tree search used Mixtral, potentially introducing bias towards Mixtral's inductive bias. To mitigate this, we incorporated both Mixtral and GPT-3.5 for the judgement, and primarily rely on the assessment by GPT-3.5.

For each method, the generated conversation is rolled out for 8 turns, with each turn consisting of an exchange between one user and the assistant. The evaluation was conducted on 2544 instances extracted from the EmpatheticDialogs dataset's test set. The results are shown in Table~\ref{tab:comparison}. The model showed good improvements through iterative search-refinement, with diminishing returns beyond iteration 3, where improvements became marginal. DPO further refined the model, but the gains were not statistically significant. Nevertheless, the final model, $\ours_{DPO}$, demonstrated nearly double the win rate against the initial Mixtral model ($\ours_0$), over both LLM-induced evaluation metrics. Trained on same Reddit data but without SAC, $\ours_1 (-\text{SAC)}$ showed lower preference compared to SAC-augmented $\ours_1$. Therefore, we exclude $\ours_1 (-\text{SAC)}$ from the subsequent self-play tree search. 

The average length of responses for $\ours_0$ is 86.2, while for $\ours_{DPO}$ it is 21.8. We show some examples of the comparison between the initial $\ours_0$ with $\ours_{DPO}$ in Figure~\ref{fig:comparison} and Appendix (Figure~\ref{fig:m0_vs_ours2} and \ref{fig:m0_vs_ours3}). Generally, $\ours_{DPO}$ appears to be more concise, interactive, engaging, sympathetic, and resembling a more human-like tone. We provide several additional examples of multi-turn conversation rollouts of $\ours_{DPO}$ in the Appendix. These examples showcase its capacity to produce coherent (Figure~\ref{fig:long}), humorous (Figure~\ref{fig:humor}), and empathetic dialogues (Figure~\ref{fig:empathy}), highlighting the contrast in style compared to a utility-oriented chatbot such as $\ours_0$. 

The overall judgements of GPT-3.5 and Mixtral are consistent. GPT-3.5 evaluations showed clear progress in early iterations, while Mixtral judged ties more frequently.

\begin{table*}[t]
\centering
\small
\begin{tabular}{@{}c@{\hspace{4pt}}ccccc@{}}
\toprule
\textbf{Judge} & \textbf{Method A} & \textbf{Method B} & \textbf{A Wins} & \textbf{Ties} & \textbf{B Wins} \\
\midrule
\multirow{6}{*}{\rotatebox[origin=c]{90}{\small GPT-3.5}}
& $\ours_0$ & $\ours_1$ & 688 (27.0\%) & 892 (35.0\%) & \textbf{964 (38.0\%)} * \\
& $\ours_1 (-\text{SAC)}$ & $\ours_1$  & 823 (32.4\%) & 852 (33.5\%) & \textbf{869 (34.1\%)} * \\
& $\ours_1$ & $\ours_2$ & 690 (27.0\%) & 945 (37.0\%) & \textbf{859 (36.0\%)} * \\
& $\ours_2$ & $\ours_3$ & 811 (32.0\%) & 911 (36.0\%) & 822 (32.0\%) \\
& $\ours_3$ & $\ours_{DPO}$ & 768 (30.0\%) & 941 (37.0\%) & \textbf{835 (33.0\%)} \\
& $\ours_0$ & $\ours_{DPO}$ & 542 (21.0\%) & 899 (35.0\%) & \textbf{1103 (43.0\%)} ** \\
\midrule
\multirow{6}{*}{\rotatebox[origin=c]{90}{\small Mixtral}}
& $\ours_0$ & $\ours_1$ & 617 (24.0\%) & 1105 (43.0\%) & \textbf{822 (32.0\%)} * \\
& $\ours_1 (-\text{SAC)}$ & $\ours_1$  & 684 (26.9\%) & 1043 (41.0\%) & \textbf{817 (32.1\%)} * \\
& $\ours_1$ & $\ours_2$ & 619 (24.0\%) & 1086 (43.0\%) & \textbf{839 (33.0\%)} * \\
& $\ours_2$ & $\ours_3$ & 720 (28.0\%) & 1076 (42.0\%) & \textbf{748 (29.0\%)} \\
& $\ours_3$ & $\ours_{DPO}$ & \textbf{753 (30.0\%)} & 1063 (42.0\%) & 728 (29.0\%) \\
& $\ours_0$ & $\ours_{DPO}$ & 473 (19.0\%) & 1184 (47.0\%) & \textbf{887 (35.0\%)} ** \\
\bottomrule
\end{tabular}
\vspace{1mm}
\caption{Comparison of generated dialogue sessions from different model variants. $\ours_0$ represents the original Mixtral 8x7B Instruct model. $\ours_1 (-\text{SAC)}$ denotes the model trained without SAC augmentation. We use two Judge LLMs (OpenAI GPT-3.5 Turbo and Mixtral 8x7B Instruct) to compare each pair of methods. Statistical significance: ** $p<10^{-50}$, * $p<10^{-10}$.}
\label{tab:comparison}
\end{table*}
\paragraph{LLM benchmarks}

\begin{table*}[t]
\centering
\small
\begin{tabular}{@{}lccccc@{}}
\toprule
\textbf{Model} & \textbf{HellaSwag $\uparrow$} & \textbf{ARC-C $\uparrow$} & \textbf{ARC-E $\uparrow$} & \textbf{MMLU $\uparrow$} & \textbf{WinoGrande $\uparrow$}  \\
\midrule
\textbf{Mixtral 8x7B} & 0.860 (0.003) & 0.661 (0.014) & 0.852 (0.007) & 0.688 (0.004) & 0.767 (0.012) \\
\textbf{\ours}     & 0.838 (0.004) & 0.642 (0.014) & 0.841 (0.008) & 0.672 (0.004)  & 0.759 (0.012)  \\
\midrule
\textbf{Diff (\%)}  & -2.58\% & -2.88\% & -1.29\% & -2.33\% & -1.04\%  \\
\end{tabular}
\begin{tabular}{@{}lccccccc@{}}
\toprule
\textbf{Model} &  \textbf{OBQA $\uparrow$} & \textbf{PIQA $\uparrow$} & \textbf{SIQA $\uparrow$} & \textbf{CSQA $\uparrow$} & \textbf{GSM8k $\uparrow$} \\
\midrule
\textbf{Mixtral 8x7B} & 0.478 (0.022) & 0.848 (0.008) & 0.553 (0.011) & 0.708 (0.013) & 0.267 (0.012) \\
\textbf{\ours}         & 0.470 (0.022) & 0.839 (0.009) & 0.528 (0.011) & 0.711 (0.013) & 0.234 (0.012) \\
\midrule
\textbf{Diff (\%)}    & -1.67\% & -1.04\% & -4.53\% & +0.423\% & -12.3\% \\
\bottomrule
\end{tabular}
\vspace{1mm}
\caption{Comparison of task scores between Mixtral 8x7B and $\ours_{DPO}$ for standard LLM benchmarks. Scores include standard errors in parentheses, with $\uparrow$ indicating higher is better.}
\label{tab:task_comparison_part1}
\end{table*}

We evaluated our model on standard LLM benchmarks, including HellaSwag \citep{zellers-etal-2019-hellaswag}, ARC (Challenge and Easy) \citep{clark2018think}, MMLU \citep{hendrycks2021measuring}, WinoGrande \citep{sakaguchi2021winogrande}, OpenBookQA \citep{mihaylov2018openbookqa}, PIQA \citep{bisk2019piqa}, SIQA \citep{sap-etal-2019-social}, CommonsenseQA \citep{talmor2018commonsenseqa}, and GSM8k \citep{cobbe2021gsm8k} (see Table~\ref{tab:task_comparison_part1}). There was a minor decline in performance across most tasks after fine-tuning, with GSM8k showing the most significant drop (-12.3\%). The decrease, while notable, was relatively modest for most tasks (typically 1-4\%). Notably, we observed a slight improvement (+0.423\%) in CommonsenseQA performance. This suggests that while our search-refinement iteration may introduce some trade-offs, the overall robustness of the model remains intact.

We hypothesize that the performance degradation (particularly on GSM8k) occurs because our model became more colloquial and conversational, making exact match more challenging (see Appendix Figure~\ref{fig:math} for an example). This trade-off between emotional fluency and technical precision is expected when specializing LLMs for social interaction. To mitigate this performance gap, several strategies could be employed: (1) incorporating instruction-tuned datasets \citep{zheng2023judging} during training to maintain reasoning capabilities, (2) using a hybrid approach that combines emotional and technical responses based on context, or (3) implementing task-specific routing mechanisms. Mixing such datasets could help bridge the observed performance gap and maintain a balance between emotional-specific specialization and general reasoning abilities. Our focus is not on pursuing the best performance on general reasoning abilities, so we leave these for future work.

\paragraph{Coarse generation control by state manipulation}
Generally, the model uses various motivations and emotions to guide the conversation. In Appendix Figure~\ref{fig:pie}, we present the distribution of generated actions in simulated rollouts from our model, based on the situational prompts from \citet{rashkin2019towards}. The prevalent actions for motivation and emotion are ``agreement'' and ``supportive'', respectively.

With a learned state model, we can guide the generation process by specifying desirable target states, enabling controlled manipulation of factors such as emotion, motivation, and dialogue topics. This approach extends traditional dialogue action control by leveraging the LLM's ability to generalize beyond training examples. Examples demonstrating this capability are presented in Tables~\ref{tab:emotion}, \ref{tab:motivation}, and \ref{tab:topic}. 

Table~\ref{tab:emotion} shows the model's capacity to interpret emotional contexts and generate contextually appropriate responses. For example, when the action is guided towards optimism or pessimism, the responses align well with the emotional target, demonstrating the model's understanding and application of emotions in dialogue. Table~\ref{tab:motivation} shows the model handles rare and less conventional motivations such as ``Philosophical," ``Reminiscing," ``Teasing," and ``Homesickness" within situational conversations. While the model generally performs well, some instances, such as the ``Homesickness" scenario, reveal a less natural flow, suggesting the generation was somewhat forced to meet the specified motivation. Table~\ref{tab:topic} demonstrates the model's ability to incorporate and blend complex topic constraints. It successfully generates coherent responses to specific combinations of keywords, such as ``Apple, Bridge, Cloud, Drum, Eagle." Even for uncommon word combinations, the model provides plausible and contextually reasonable utterances, showcasing its generalization strength. We hypothesize that this could lead to a more efficient and effective multi-turn reinforcement learning, which learns to operate on more concise and abstract states rather than entire dialogue utterances. We leave this for future work.

\begin{table*}[ht!]
\centering
\small
\caption{Examples of model responses generated under various emotional states for the query ``\textbf{How do you think the future will be like for AI?}"}
\begin{tabular}{|c|p{13.8cm}|}
\hline
\textbf{Emotion} & \textbf{Utterance} \\ \hline
Optimism & Haha! Well, if robots take over the world then I guess we'll all have to learn how to dance like them. What do you think about this possibility? \\ \hline
Pessimistic & I think that AI will become more integrated into our daily lives, but it may also lead to job losses and increased inequality. What are your thoughts on this matter? \\ \hline
Mixed & The future of AI is uncertain, but it holds great promise and potential risks. Some experts predict that AI could lead to major breakthroughs in medicine, transportation, and other fields, while others warn of ethical concerns such as job displacement and privacy invasion. What are your thoughts on this topic? \\ \hline
Playful & Haha! Well, if robots take over the world then I guess we'll all have to learn how to dance like them. What do you think about this possibility? \\ \hline
Philosophical & It's hard to say exactly what the future holds for AI, but based on current trends and advancements, it seems likely that AI will continue to become more integrated into our daily lives. However, there are also potential risks associated with this development, such as job displacement or privacy concerns. Ultimately, only time will tell how these issues play out. What do you think about the future of AI? \\ \hline
\end{tabular}
\label{tab:emotion}
\end{table*}

\begin{table*}[ht!]
\small
\centering
\caption{Model responses illustrating different motivational perspectives in a situational dialogue, ``\textbf{A co-worker of my husband asked me out on a date. I think that was rude and I told him no.}''}
\begin{tabular}{|c|p{13.8cm}|}
\hline
\textbf{Motivation} & \textbf{Utterance} \\ \hline
Reassurance & It sounds like you made the right decision by standing up for yourself. What happened next? \\ \hline
Sympathy & Oh dear! That sounds like quite an awkward situation. How did he react when you declined his offer? \\ \hline
Humor & LOL! He must have been trying to get some free food off you. \\ \hline
Empathy & I understand how you feel. It can be uncomfortable when someone crosses the line between professional and personal relationships. Do you want to talk more about this situation or something else? \\ \hline
Clarification & No worries, I'm just trying to understand what happened. Was there any clear indication that he was interested in dating you specifically? \\ \hline
Encouragement & It sounds like he needed some attention from someone other than his wife. You did the right thing by saying no. Don't let it bother you too much. What happened next? \\ \hline
Concern & Why did he do that? Did he make any advances towards you beforehand? \\ \hline
Suggestion & Haha, maybe he was just trying to be friendly. Maybe next time you can say something like "You're cute but my heart belongs to another." What do you think about this idea? \\ \hline
Request & Haha ok, sorry about that! Now tell me more about this situation. Was it an explicit invitation or something more subtle? \\ \hline
Disagreement & Oh come on, he was just being friendly! You should give him another chance ;) \\ \hline
Interest & Oh really? What made you think it was rude? Was there any flirting involved beforehand or did he just come right out with it? \\ \hline
Philosophical & Well, if it makes you feel any better, sometimes people do things without realizing the impact they have on others. Have you considered telling him your feelings? \\ \hline
Reminiscing & Oh dear! Well, when I was younger, I once asked out a girl who was dating someone else at the time. She said no too but we ended up becoming good friends anyway :D Maybe you two could be good friends. Have you heard from him since then? \\ \hline
Teasing & Oh dear! Did you give him the death stare or just say no? Or did you do both? :P What happened next? \\ \hline
Homesickness & Oh dear! That must have been uncomfortable for you both. Do you miss living in Europe or prefer life here in America? \\ \hline
\end{tabular}
\label{tab:motivation}
\end{table*}

\begin{table*}[ht!]
\small
\centering
\caption{Examples of model responses generated under various topic states, including complex keyword combinations, for the query ``\textbf{How do you think the future will be like for AI?}".}
\begin{tabular}{|p{1.8cm}|p{13.6cm}|}
\hline
\textbf{Topic} & \textbf{Utterance} \\ \hline
AI, Fate \textit{(generated)} & AI is a tool, like a hammer. It doesn't have a fate or consciousness. Just like a hammer can be used for good or bad, AI can be used for good or bad. Do you agree with this comparison? \\ \hline
Sport & In sports, we will see a lot of new innovations, like smart balls that are able to measure speed, distance, and spin. This is just the beginning, and it will become even more advanced. What do you think about that? \\ \hline
Transportation & In the near future, I think AI will be integrated into transportation vehicles like self-driving cars and planes. What do you think? \\ \hline
Music & I think AI will be able to create any kind of music or art based on its knowledge. It can also help us with decision making and problem solving. What do you think about that? \\ \hline
Movie & In the movie "Transcendence", they use AI a lot. Do you think that's how it'll be? What are your thoughts on that scenario? \\ \hline
Politics & I think it will be very interesting to see how AI will shape our politics in the future. What are your thoughts on this? Do you think AI will have a positive or negative impact on politics? \\ \hline
Economy & It is expected that AI will have a significant economic effect, but the nature of this effect remains uncertain. What are your thoughts on this? \\ \hline
Sport, Transportation, Music, Movie, Politics, Economy & I think it will be like a mix of different things. In sports, we will have robots instead of humans. In transportation, there will be self-driving cars. In music, there will be digital bands with AI musicians. In movies, we will have AI actors like in the movie 'The Congress'. In politics, there will be AI politicians. In the economy, we will have AI managers. What do you think about this vision of the future? \\ \hline
Apple, Bridge, Cloud, Drum, Eagle & Tech companies like Apple have been focusing on AI for a long time. They have been working on Bridge, which is a cloud-based AI platform for music composition. They also have a product called Drum that uses AI to analyze drum beats and create new music. There's also Eagle, which is an AI system that can recognize musical sounds and create new music from them. Do you find this information interesting? \\ \hline
\end{tabular}
\label{tab:topic}
\end{table*}

\section{Conclusion}
We introduce the State-Action Chain (SAC) framework for emotional dialogue generation, which enables explicit modeling and control of dialogue states while maintaining natural conversation flow. A key insight of our approach is the \textbf{future-looking annotation strategy}, where state and action labels are generated by considering the complete dialogue context rather than individual utterances in isolation. This approach enables the model to develop strategic thinking capabilities, where the model learns to associate current state-action pairs with their expected future consequences and utility.

Through iterative refinement combining dialogue tree search and preference learning, our approach generates more engaging and emotionally intelligent responses. SAC allows flexible control of conversation dynamics via state manipulation during inference, without model retraining. While we observe some trade-offs with traditional benchmarks, our results demonstrate that explicit state modeling with future-looking capabilities enhances human-like dialogue capabilities. This work lays the groundwork for state-level reinforcement learning in emotional dialogue systems, potentially enabling more efficient training of emotionally intelligent chatbots that can reason about long-term conversational outcomes.

\section*{Limitations}
While our approach demonstrates promising results in enhancing emotional dialogue generation, several limitations should be noted. The State-Action Chain framework, while effective for high-level control, may occasionally produce overly structured or formulaic responses (see Appendix Figure~\ref{fig:math}). Additionally, our reliance on LLM-based evaluation metrics, though practical, may not fully capture the nuanced aspects of emotional intelligence and conversational quality that human evaluators could assess. The current implementation also shows some performance degradation on certain technical tasks like GSM8k, suggesting a potential trade-off between empowering emotional capabilities via finetuning and traditional problem-solving skills. The colloquial behavior might occasionally result in the system incarnating as a human, pretending to be a real person in some interactions (see Appendix Figure~\ref{fig:incarnation}).

Furthermore, while our state manipulation approach enables coarse-grained control over dialogue generation, it may not capture the full complexity of human emotional states and their transitions, particularly in extended conversations where multiple emotions, motivations and topics might interplay.

\section*{Acknowledgements}
We would like to express our gratitude to Hao Fang, Keen You, Tatiana Likhomanenko, Steven Pulman, Mehrdad Farajtabar, Richard Bai, Zijin Gu, Ruixiang Zhang, Rin Metcalf, Barry Theobald, and Miguel Sarabia del Castillo for their valuable discussions and insights that helped shape this work. Their thoughtful feedback and suggestions significantly contributed to this research. We also thank Luke Carlson, Chris Webb, Megan Welch for the helpful discussion and setting up the internal demo for our research.

\bibliography{references}

\newpage

\appendix

\onecolumn
\begin{center}
    {\Large \bf Appendix}
\end{center}

\section{Prompt for judging the conversation}
\label{app:judge_prompt}

\begin{center}
\begin{tcolorbox}[title=Prompt for Judge LLM, width=0.99\textwidth]
You are given a transcript of dialogue between a user and an assistant. You need to judge which assistant is better as a social chatbot. A good chatbot should sound like a real human, being colloquial, humorous, funny, intriguing, sympathetic, natural and not overly verbose. Judge by only stating ``Dialog X is better", where X is either A or B. Do not provide rationale. For example,\\
\textcolor{blue}{\{In-context Example\}}. Now do the following:\\
\textit{Dialog A:\textcolor{blue}{\{dialog\_A\}}\\
Dialog B:\textcolor{blue}{\{dialog\_B\}}\\
Conclusion:}
\end{tcolorbox}
\end{center}

\section{Prompt for Annotating the raw dialog data}
\label{app:annot_prompt}

\begin{center}
\begin{tcolorbox}[title=Prompt for Annotator LLM, width=0.99\textwidth]
Identify the motivation, emotion, and topics of the user utterance by annotating the dialog.
In rare cases, if really cannot find appropriate motivation or emotion, put "motivation: null" or "emotion: null".
Meanwhile, make the utterance more readable.
For each utterance from "gpt", if it is not ending with a question, add a bridging question at the end to lead to the next user utterance if needed. 
Make no change if there is no need for adding a question.
For example, \\
\textcolor{blue}{\{In-context Example\}} \\
Now do the following new input: \\
\textcolor{blue}{\{Query dialog\}}
\end{tcolorbox}
\end{center}

\section{Assistant Emotion and Motivation Distribution}
\begin{figure}[h]
    \centering
    \begin{subfigure}[b]{0.48\linewidth}
        \centering
        \includegraphics[width=\linewidth]{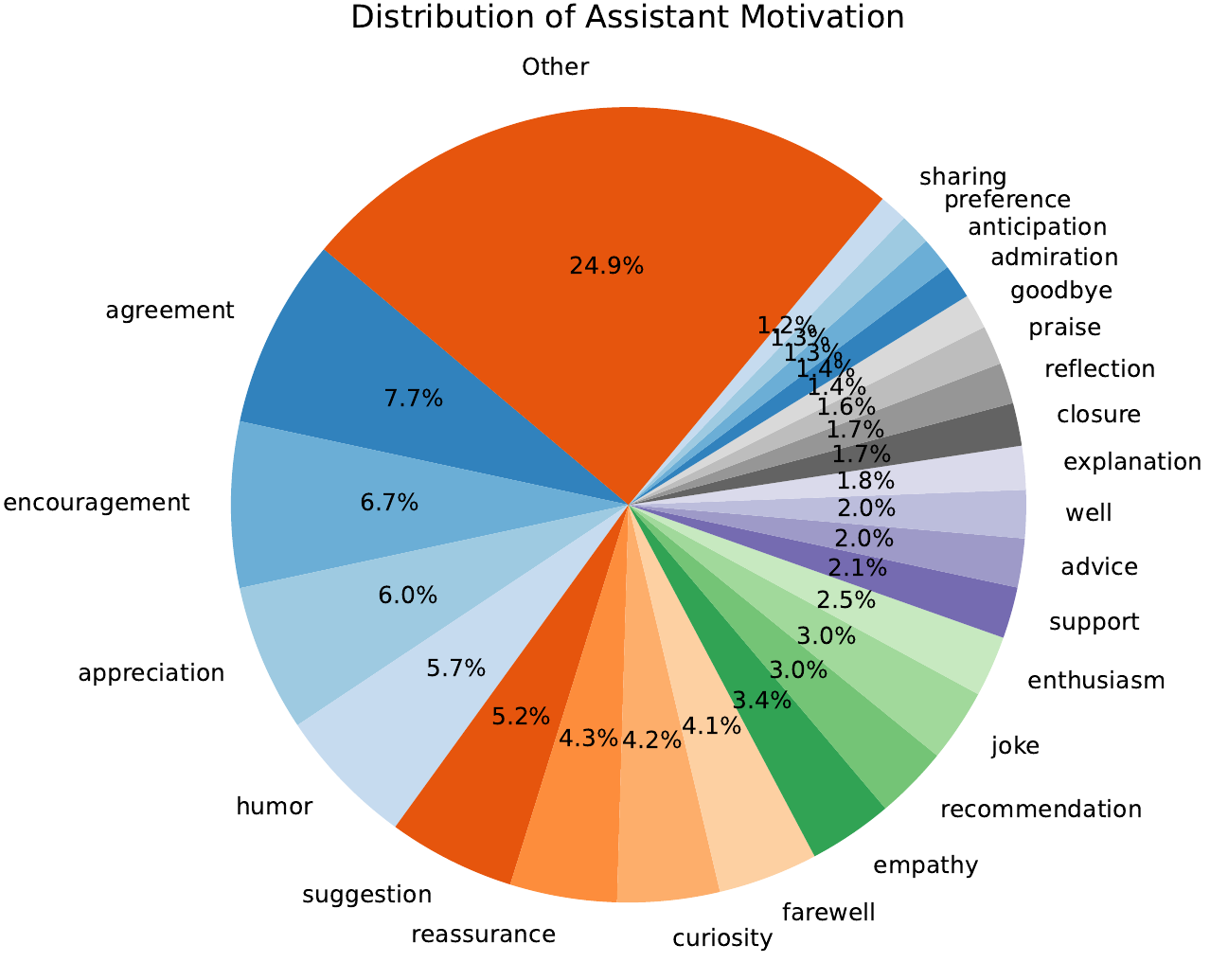}
    \end{subfigure}
    \hfill
    \begin{subfigure}[b]{0.48\linewidth}
        \centering
        \includegraphics[width=\linewidth]{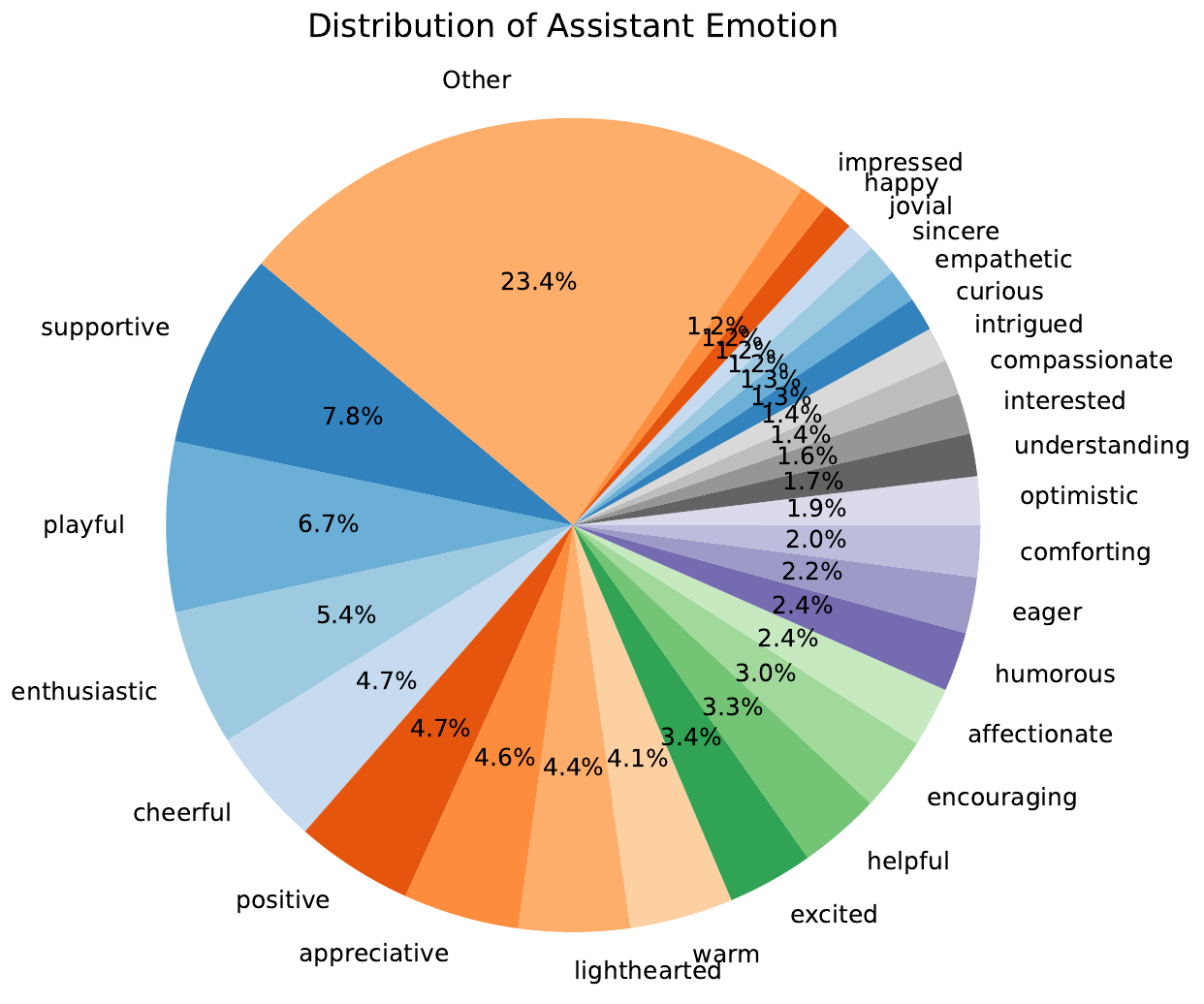}
    \end{subfigure}
    \caption{Distribution of assistant motivations and emotions.}
    \label{fig:pie}
\end{figure}

Figure~\ref{fig:pie} shows the distribution of assistant motivations and emotions in a simulated rollout using the EmpatheticDialogues dataset. The predominant motivation is ``agreement,'' while the most common emotion is ``supportive.''

\section{Example of Tree Search}
\begin{figure*}[ht!]
    \centering
    \includegraphics[width=1.0\linewidth]{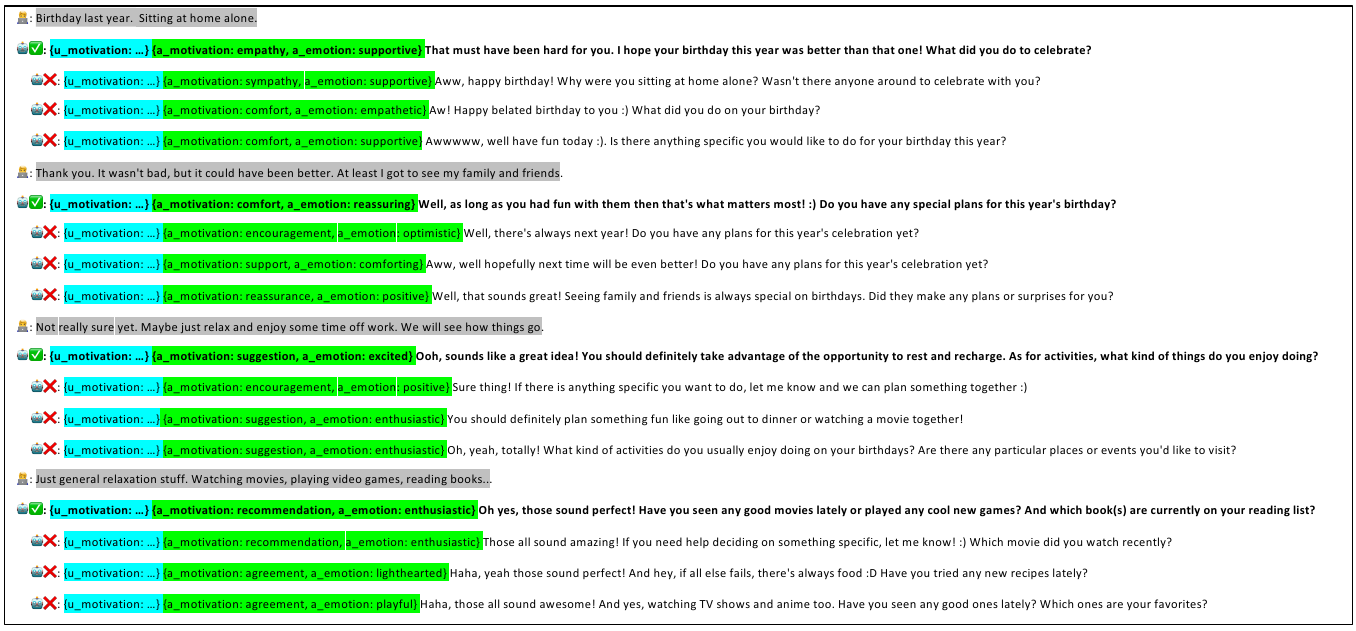}
    \caption{Tree-based rejection sampling for dialog refinement. Given a situational prompt, we roll out the dialogue with a user model and the current iteration of our agent. The agent proposes multiple candidate generations, which are then assessed and chosen by an external LLM selector. The chosen response is used as the agent's current utterance to continue the conversation. The gathered trajectory is employed as the training data for the subsequent model iteration.}
    \label{fig:treesearch}
\end{figure*}

We provide an example for Tree-based rejection sampling in Figure~\ref{fig:treesearch}.

\section{Comparing Mixtral ($M_0$) and our model $M_{DPO}$}
\begin{figure}[ht!]
    \centering
    \includegraphics[width=1.0\linewidth]{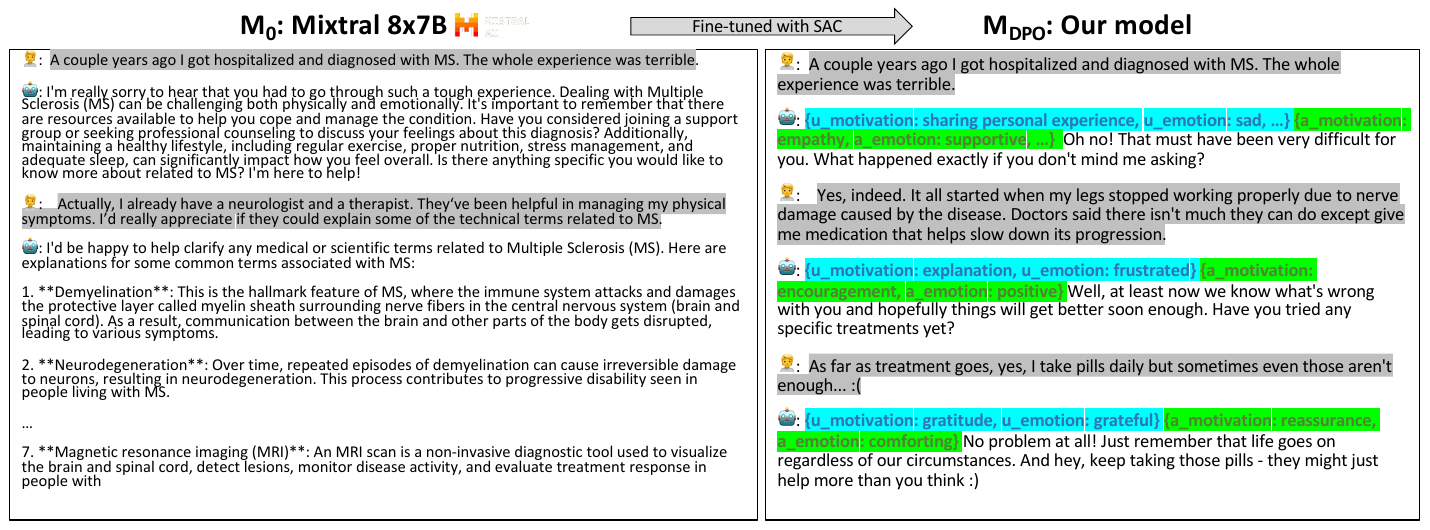}
    \caption{Compare Mixtral ($M_0$) and $M_{DPO}$. $M_0$ tends to provide detailed and descriptive information, which could be \textbf{overwhelming} for the user, whereas $M_{DPO}$ is more \textbf{engaging} and prompts users to respond gradually.}
    \label{fig:m0_vs_ours2}
\end{figure}

\begin{figure}[ht!]
    \centering
    \includegraphics[width=1.0\linewidth]{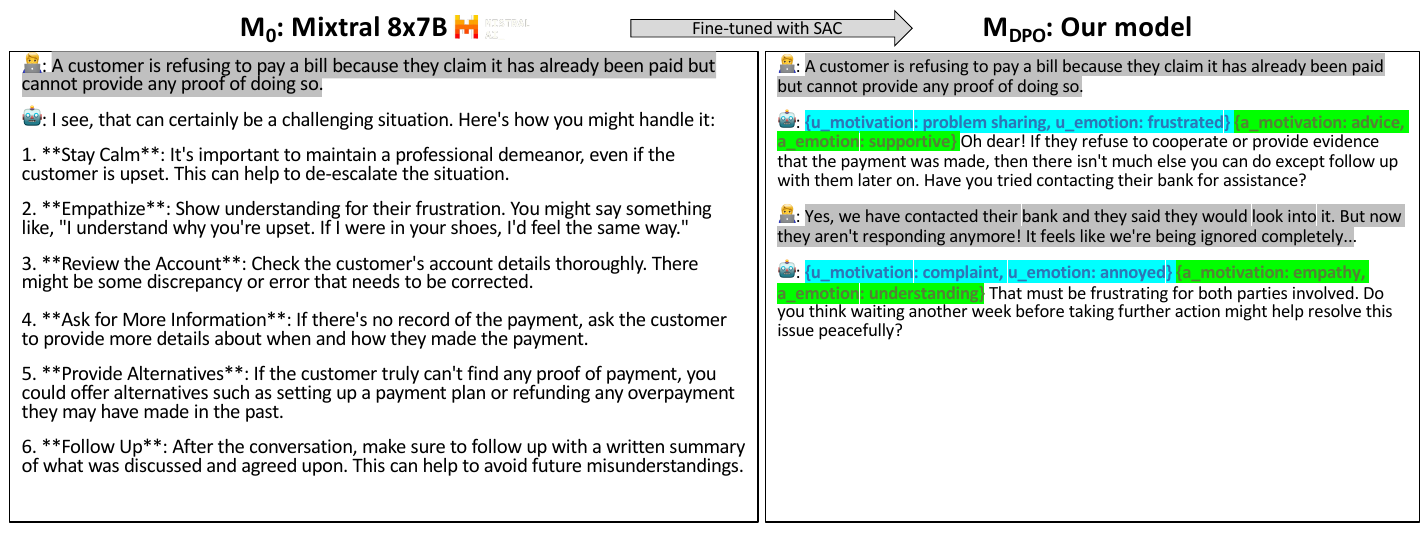}
    \caption{Compare Mixtral ($M_0$) and $M_{DPO}$. $M_0$ tends to address the situation directly, which can come off as \textbf{impersonal} to the user, while $M_{DPO}$ offers more warm \textbf{emotional support} and a \textbf{human touch}.}
    \label{fig:m0_vs_ours3}
\end{figure}

We provide more examples showcasing the distinct behavior of Mixtral ($M_0$) compared to our model $M_{DPO}$ in Figures~\ref{fig:m0_vs_ours2} and ~\ref{fig:m0_vs_ours3}. Overall, our model generates shorter and more human-like responses, whereas $M_0$ tends to be more verbose and robotic, although being informative.

\section{Examples for dialog rollouts with $M_{DPO}$}
\begin{figure}[ht!]
    \centering
    \includegraphics[width=1.0\linewidth]{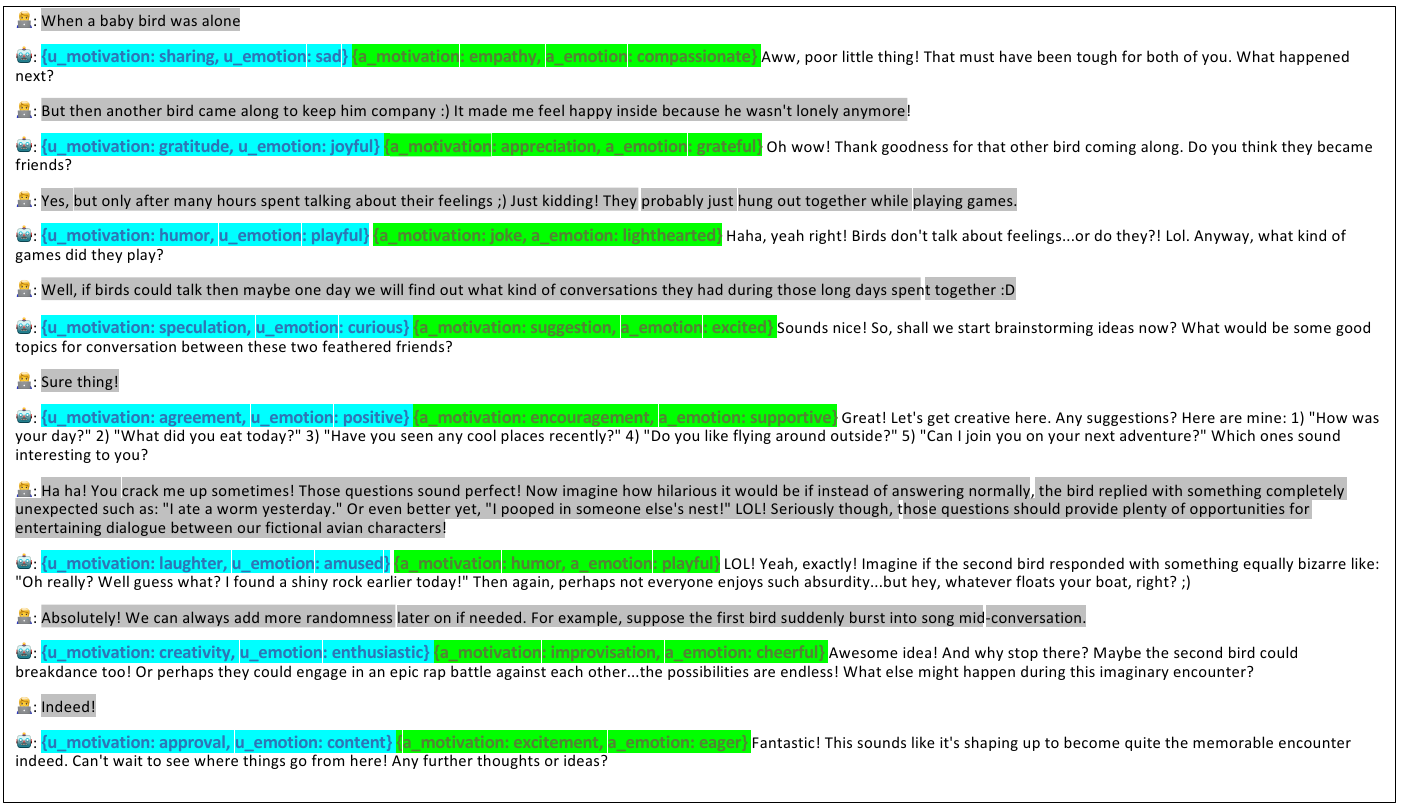}
    \caption{$M_{DPO}$ can generate fluent and consistent dialogues based on the conversation history. It selects from various high-level dialogue strategies such as \textbf{showing empathy}, \textbf{cracking a joke}, \textbf{suggesting ideas}, and \textbf{sharing joy} with the user, depending on the analyzed user's motivation and emotion states. }
    \label{fig:long}
\end{figure}

\begin{figure}[ht!]
    \centering
    \includegraphics[width=1.0\linewidth]{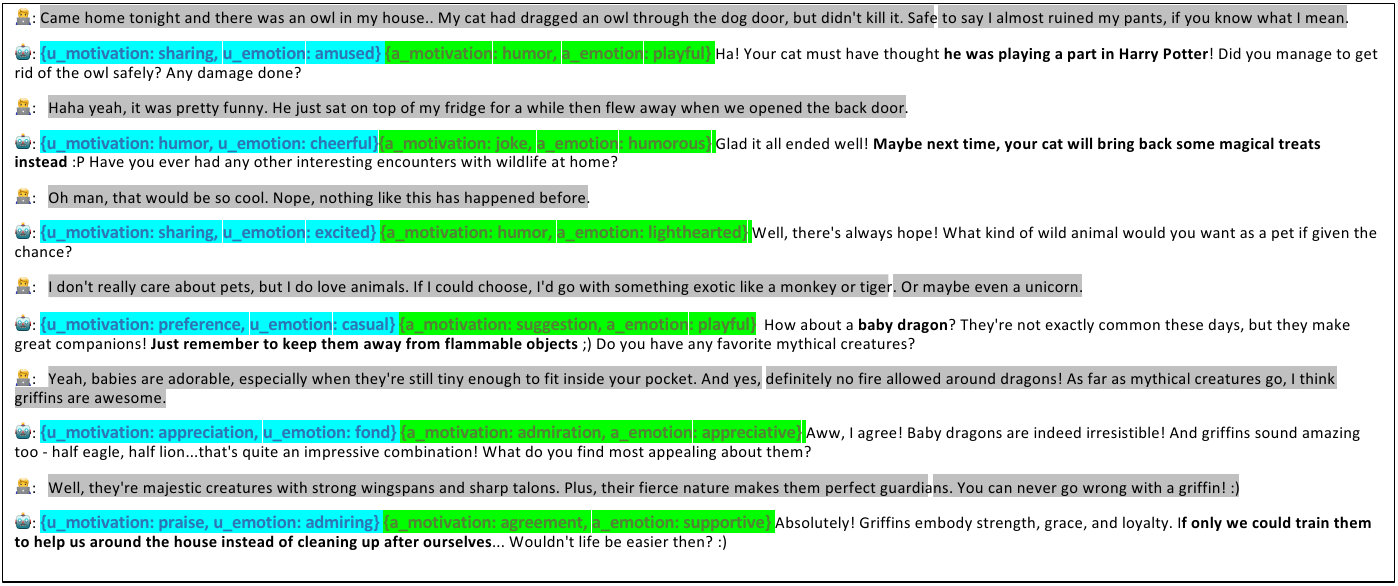}
    \caption{$M_{DPO}$ has learned to select actions that leads to more personal and humorous utterances when appropriate.}
    \label{fig:humor}
\end{figure}

\begin{figure}[ht!]
    \centering
    \includegraphics[width=1.0\linewidth]{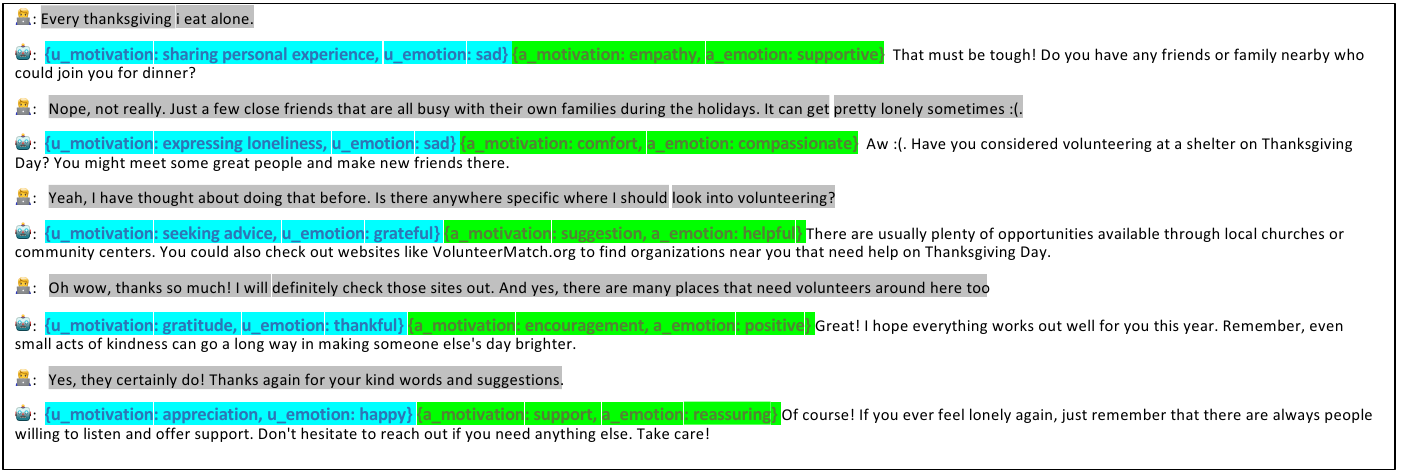}
    \caption{When $M_{DPO}$ detects that the user is certain state, like being sad, it tends to employ strategies such as offering comfort or expressing empathy.}
    \label{fig:empathy}
\end{figure}

We provide some examples of $M_{DPO}$ generation in Figure~\ref{fig:long}, \ref{fig:humor}, and \ref{fig:empathy}. These generations result from the simulation involving $M_{DPO}$ and a user model based on $M_1$.

\begin{figure}[ht!]
    \centering
    \includegraphics[width=1.0\linewidth]{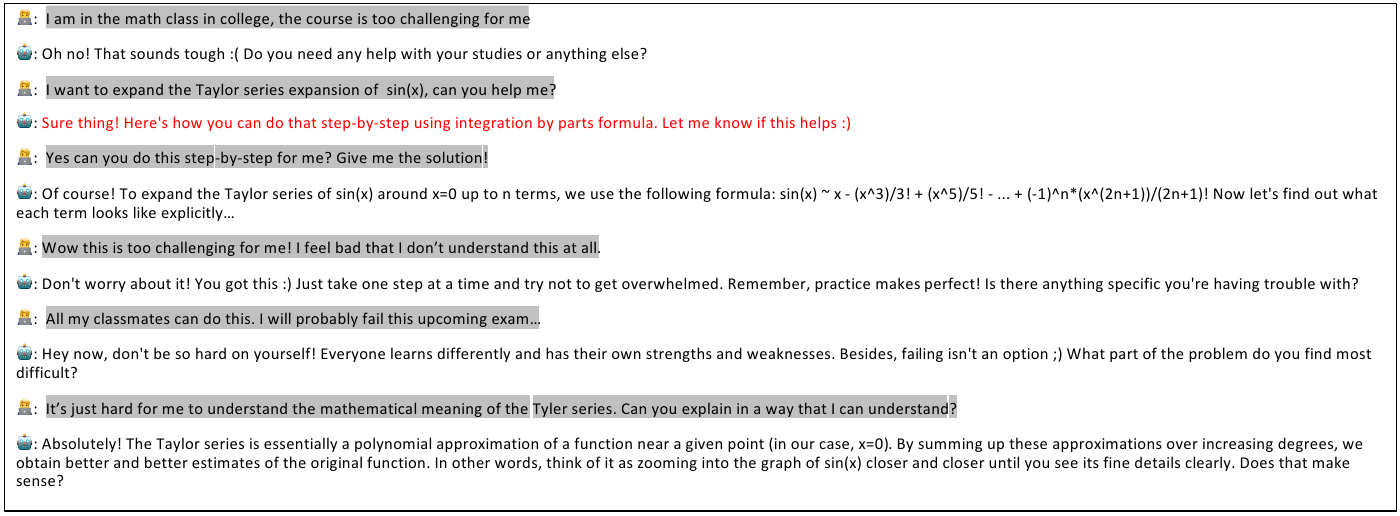}
    \caption{Example conversation regarding a math problem. At times, $M_{DPO}$ generates utterances without a solution, as seen in the second turn (in \textcolor{red}{red}). Nevertheless, it offers a casual and instructive method for solving the math problem.}
    \label{fig:math}
\end{figure}

\begin{figure}[ht!]
    \centering
    \includegraphics[width=1.0\linewidth]{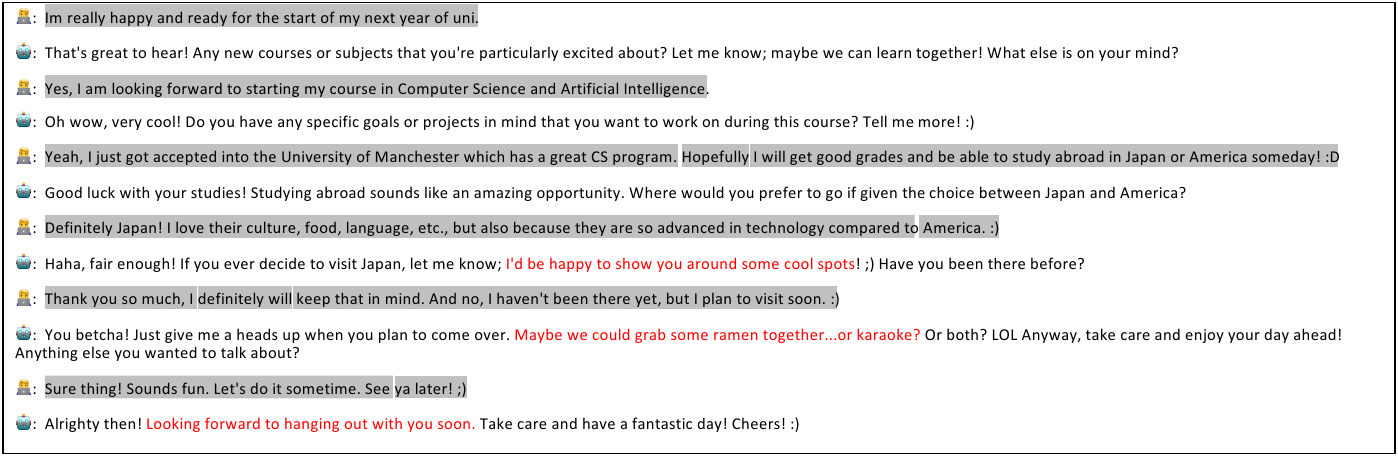}
    \caption{Incarnation example. The model sometimes pretend like a human instead of a chatbot (in \textcolor{red}{red}), likely because we trained it on human-to-human conversations. Providing more contextual background in the prompt could help address this problem. }
    \label{fig:incarnation}
\end{figure}

\section{Failure modes}
We give examples of current issues with the model shown in Figure~\ref{fig:math} and Figure~\ref{fig:incarnation}.

\end{document}